\pdfoutput=1
\documentclass{article}

\usepackage[utf8]{inputenc}
\usepackage{amsmath}
\usepackage[justification=centering]{caption}
\usepackage{subcaption}
\usepackage{graphicx}
\usepackage{booktabs}
\usepackage{makecell}
\usepackage{multirow}

\usepackage[final]{corl_2018} % Uncomment for the camera-ready ``final'' version

\title{Leveraging Deep Visual Descriptors\\ for Hierarchical Efficient Localization}

\author{
 Paul-Edouard Sarlin \qquad Frédéric Debraine \qquad Marcin Dymczyk \\ \textbf{Roland Siegwart \qquad Cesar Cadena} \\[2mm]
 Autonomous Systems Lab, ETH Zürich\\
}

\begin{document}
\maketitle

%===============================================================================

\begin{abstract}
Many robotics applications require precise pose estimates despite operating in large and changing environments. This can be addressed by visual localization, using a pre-computed 3D model of the surroundings. The pose estimation then amounts to finding correspondences between 2D keypoints in a query image and 3D points in the model using local descriptors. However, computational power is often limited on robotic platforms, making this task challenging in large-scale environments. Binary feature descriptors significantly speed up this 2D-3D matching, and have become popular in the robotics community, but also strongly impair the robustness to perceptual aliasing and changes in viewpoint, illumination and scene structure.
In this work, we propose to leverage recent advances in deep learning to perform an efficient hierarchical localization. We first localize at the map level using learned image-wide global descriptors, and subsequently estimate a precise pose from 2D-3D matches computed in the candidate places only. This restricts the local search and thus allows to efficiently exploit powerful non-binary descriptors usually dismissed on resource-constrained devices.~Our approach results in state-of-the-art localization performance while running in real-time on a popular mobile platform, enabling new prospects for robotics research.\footnote{Code and video available at \url{http://github.com/ethz-asl/hierarchical_loc}\vspace{-1ex}}
\end{abstract}


%===============================================================================
\section{Introduction}
% State the problem
An increasing number of mobile robotics applications are deployed in large, uncontrolled, and GPS-denied environments, such as for city-wide parcel delivery or industrial plant inspection. Tasks like path planning and object avoidance rely on accurate high-frequency 6 degree-of-freedom (6-DoF) pose estimates with respect to a prior map. This is commonly achieved with image-based localization and a~global 3D model composed of sparse visual keypoints obtained from a Simultaneous localization and mapping (SLAM) pipeline.

% Why it's challenging
A critical step of the localization pipeline is the computation of correspondences between 2D local features in the query image and 3D points in the map. While this is often accomplished with a nearest neighbor search in a common descriptor space, city-block-scale maps contain a large number of high-dimensional descriptors, making this search intractable on computationally-limited platforms, which are often used in robotics. Additionally, as the maps increase in size, ambiguity arises and significantly reduces the descriptors discriminability.

% Common solutions: cheaper direct matching or retrieval systems
The robotics community has developed a wide range of methods to make this 2D-3D matching more efficient. State-of-the-art localization methods rely on a heavily-optimized direct matching of binary local descriptors, which, however, further impair the robustness to perceptual aliasing and changes of illumination, viewpoint, and structure. Advances in convolutional neural networks (CNN) have largely improved the performance of image retrieval systems, but these are still not able to directly estimate a centimeter-precise pose, and are difficult to deploy on resource-constrained devices.

% Our proposal
In this paper, we propose to tackle the visual localization problem in a hierarchical manner, leveraging global learned descriptors and local powerful handcrafted features for a coarse-to-fine global-to-local search resulting in an accurate pose estimate. Our work shows how exploiting recent neural models and knowledge compression techniques can drastically boost the performance of classical localization systems while preserving run-time efficiency on mobile devices. Overall, our contributions are as follow:
\vspace{-0.2cm}
\begin{itemize}
    \item We propose a simple, yet efficient, scheme to compress a large image retrieval model into a smaller network with mobile real-time inference capability;
    \item We show how this learned prior can significantly improve the localization performance by deriving a hierarchical search;
    \item We demonstrate that our proposed approach outperforms current state-of-the-art robotic relocalization systems and particularly emphasize its tractable compute requirements.
\end{itemize}

\begin{figure}[!htb]
    \centering
    \captionsetup{justification=justified}
    \includegraphics[width=\linewidth]{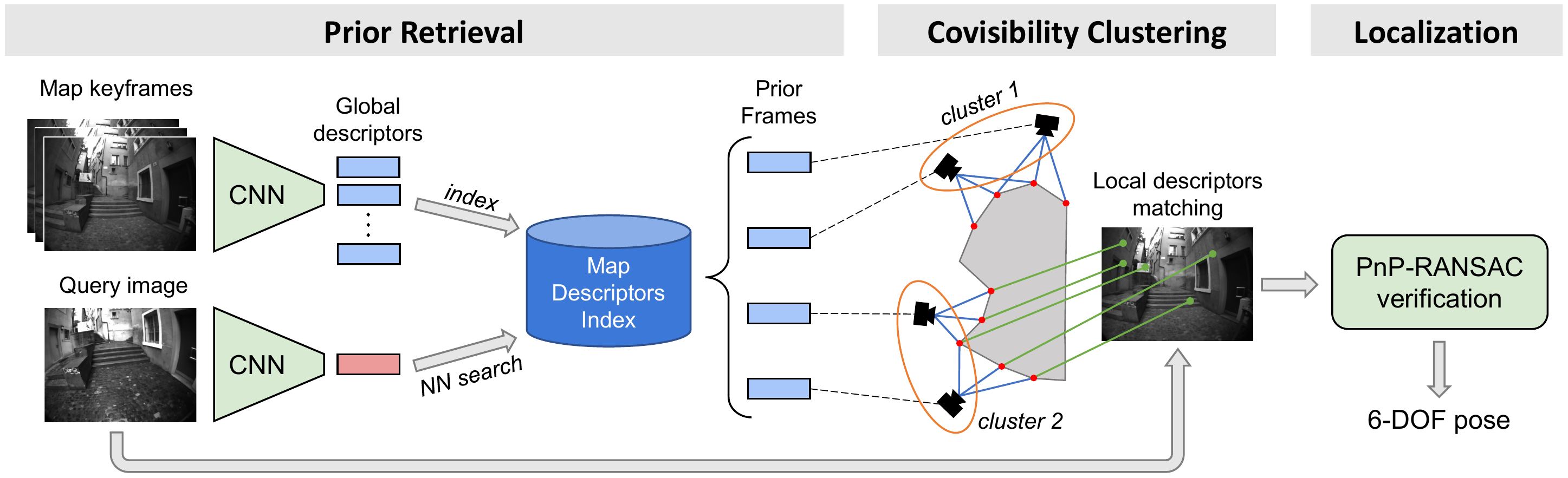}
    \caption{\textbf{Overview of our hierarchical localization system.} For a given query image, a coarse search first finds candidate keyframes in the map using global descriptors learned by a CNN. These prior frames are clustered into places, and a local search over expensive local descriptors is performed for each of them until a valid 6-DoF pose is estimated.\label{fig:full_pipeline}}%
\end{figure}%
\vspace{-1.5ex}

%===============================================================================
\section{Related Work}
\label{sec:rel-work}

In its early days, visual localization has been cast as a place recognition task, whose goal is to categorize a given query image into a limited set of discrete places. This has traditionally been studied as an image retrieval problem, where the location of the query is approximated from the locations of visually similar images retrieved from a database. Each of these is associated with a global descriptor, often defined as the aggregation of handcrafted local descriptors, like SIFT~\citep{sift}, using techniques such as bag-of-visual-words~\citep{fabmap, dbow2} or VLAD~\citep{vlad}. Recent advances in deep learning and the availability of large labelled visual datasets have enabled powerful end-to-end models, such as DELF~\citep{delf} or NetVLAD~\citep{netvlad}, to considerably improve the retrieval performance. These can however only estimate a coarse location whose accuracy is limited by the place discretization.

Many robotic applications require a precise 6-DoF pose, thus motivating the development of novel methods based on local feature matching. Given a pre-computed 3D point cloud representing the scene, \citet{irschara} proposed to find 2D correspondences between the query and similar images retrieved from a database. While this is fast, \citet{sattler11} later showed that directly matching 2D features in the query with 3D points achieves substantially better results. Yet performing the 2D-3D matching with such expensive descriptors is intractable on mobile and robotics platforms. In order to make this search more efficient, several solutions have been proposed, such as vocabulary trees~\citep{vocabulary-tree} and prioritized search~\citep{prioritized-search}. \citet{middelberg} proposed to offload the main computational burden of the localization to a remote server, which is however not possible in many robotic scenarios, motivating further research on search optimization. This yielded the emergence of binary descriptors such as BRISK~\citep{brisk} and FREAK~\citep{freak}, whose Hamming distance is fast to compute, and whose matching can be significantly sped-up using a inverted multi-index and product quantization~\citep{get-out-of-my-lab}, resulting in current state-of-the-art performance on resource-constrained systems.

Despite their wide adoption~\citep{orb-slam}, binary descriptors still suffer from a lack of discriminability and their matching often fails in highly repetitive environments. To address these issues, \citet{learned-proj} learn a projection that improves the matchability and invariance of these descriptors. \citet{experience-based} individually match the query against separate experiences that are perceptually different, thus hinting that a higher level view of the scene appearance is useful to reduce the size of the problem. The performance of these approaches is however limited by the intrinsic fragility of the underlying efficient descriptors. While learned local features have recently shown promising results~\citep{superpoint, sips}, we demonstrate here that expensive hand-crafted descriptors can already provide a significant, yet tractable, performance improvement when preceded by a global search.

With the advent of deep learning, some methods attempt to directly regress a full 6-DoF pose using deep networks. They however either require training in the target environment~\citep{posenet}, or work on image sequences~\citep{valada-regression}, and overall do not achieve the high accuracy required by robotic navigation and control, nor do they work on 3D models built with feature-based pipelines. In this work, we propose to combine the two research directions of image retrieval and 2D-3D direct matching for a fast, robust and accurate localization. Inspired by the performance of LocalSfM from \citet{loc-benchmark}, we leverage deep networks developed for place recognition to estimate candidate places, and rely on expensive hand-crafted local features to estimate a pose. Unlike previous work on learning-based localization, our method 1) estimates a centimeter-precise 6-DoF pose, 2) works with existing 3D maps, 3) does not require but can benefit from training in the target environment, and 4) runs in real-time on a mobile platform.

%===============================================================================
\section{Hierarchical Localization Method}
\label{sec:method}

We aim at restricting the search space of local 2D-3D matches by reducing the number of 3D points considered. We employ a hierarchical approach similar to how humans naturally localize in a previously visited environment by first looking at the global scene appearance and subsequently inferring an accurate location from a set of likely places using local visual clues. The process is shown in Figure~\ref{fig:full_pipeline}. In a first step, a CNN-based image retrieval system infers a set of database prior frames that are most likely to correspond to the same location as the query image. Candidate places are then determined using the covisibility graph of the map, and an iterative matching-estimation process eventually computes the exact pose.

\subsection{Prior retrieval}
Global feature descriptors are first estimated and indexed for each keyframe used to build the map. For each query image, its corresponding descriptor is computed and its nearest neighbors are retrieved from the index.

\begin{figure}[!htb]
    \centering
    \captionsetup{justification=justified}
    \includegraphics[width=0.8\linewidth]{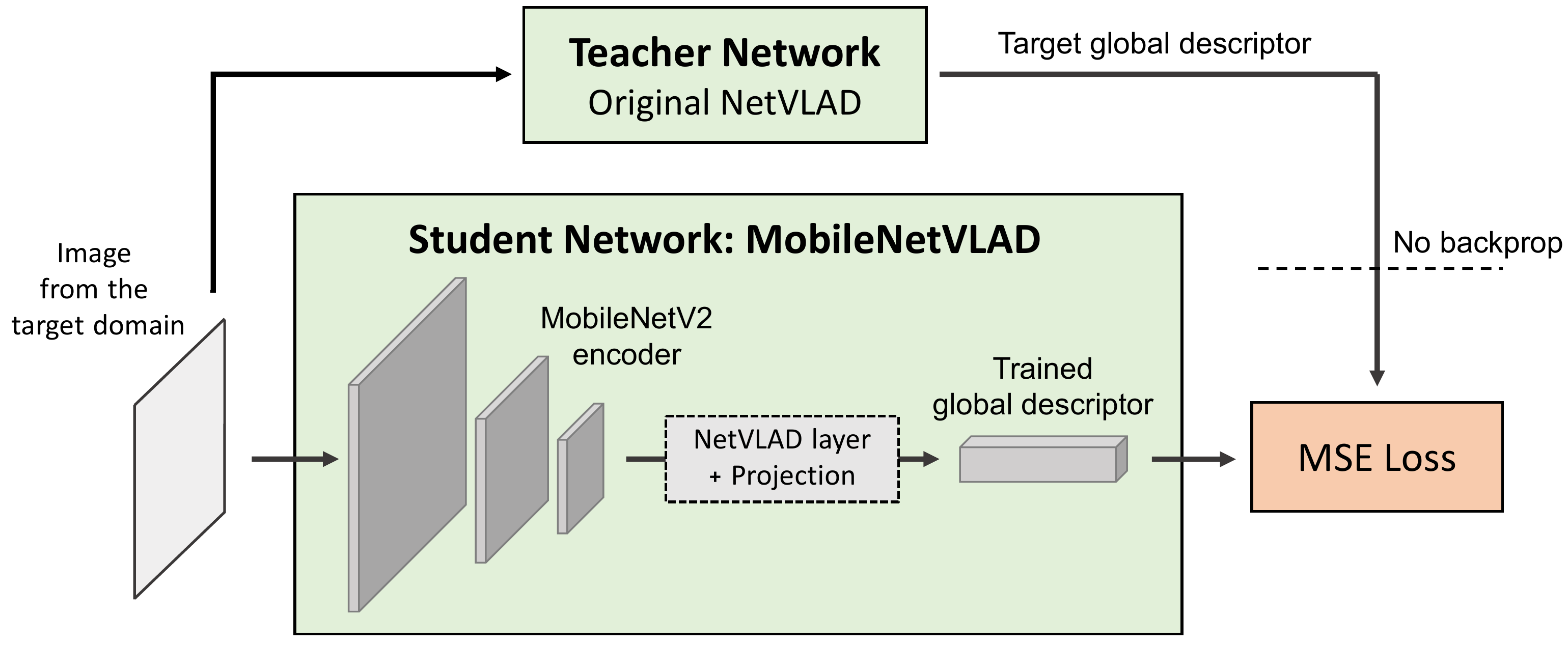}
    \caption{\textbf{Training process of MobileNetVLAD.} We use knowledge distillation~\citep{distillation} to train the student by minimizing the Mean Squared Error (MSE) loss between its predictions and target global descriptors predicted by a trained NetVLAD network (the teacher). The training data does not require labels, such as the triplets that the teacher uses, and can be restricted to a specific environment to bias the generalization of the new model in a target domain.\label{fig:mobilenetvlad}}%
\end{figure}

Following recent advances in image retrieval, a CNN predicts the global descriptor of a given image in a two-step manner: an encoder first computes a high-level feature map, which is subsequently aggregated into a single global descriptor. While channel-wise global max-pooling is a trivial aggregation, the VLAD layer~\citep{netvlad} has recently shown impressive results in the image retrieval task. The VGG-16~\citep{vgg} encoder used by the original NetVLAD network is good at learning useful representations but too computationally expensive to be deployed on a mobile platform. Recent work~\citep{mobilenet} introduced MobileNet, an encoder architectures tailored for resource-constrained devices. As the training procedure of NetVLAD is complex, and no mobile-friendly variant has yet been proposed, we use knowledge distillation~\citep{distillation} to train a smaller network (the student) at learning global descriptors predicted by NetVLAD (the teacher). As a pre-trained NetVLAD has been released by \citet{netvlad}, this drastically reduces the training time of the new model and improves its generalization. The training process is shown in Figure~\ref{fig:mobilenetvlad}. We design the student network as a scaled-down version of NetVLAD using a MobileNet encoder followed by a smaller VLAD layer, and name it MobileNetVLAD. A learned linear fully-connected layer down-projects the student descriptors so that their dimension matches with the target descriptors. 

NetVLAD is originally trained with a triplet loss, which minimizes the distance between descriptors of images corresponding to the same location. By construction, descriptors predicted by MobileNetVLAD exhibit the same properties. We thus obtain coarse estimates of the query location by retrieving a fixed number of its neighboring keyframes in the descriptor space using a k-d tree. In order to facilitate this search, the descriptor dimension is reduced using PCA, which can be computed on the indexed images, yielding a more homogeneous descriptor distribution in image domains significantly different from the training dataset. The projected descriptor is eventually L2-normalized.

The original NetVLAD model was trained on images of very different appearances, e.g.\ changing weather conditions, and has learned to focus on time-invariant visual clues at a high structural level. This prior knowledge of perceptual changes is then transferred to MobileNetVLAD through distillation. We thus call the retrieved images \emph{prior frames}.

\subsection{Covisibility Clustering}
All the retrieved prior frames may not correspond to the same area in the map because of perceptual aliasing. They are however clustered in \emph{places}, i.e. distinct locations, as subsequent keyframes have common visual clues and are thus likely to be retrieved together. Performing the local search on each of them independently makes it more robust, as it reduces the number of wrong local matches in the correct place, thus increasing the chance of a successful localization.

\begin{figure}[!htb]
    \centering
    \captionsetup{justification=justified}
    \includegraphics[width=0.8\linewidth]{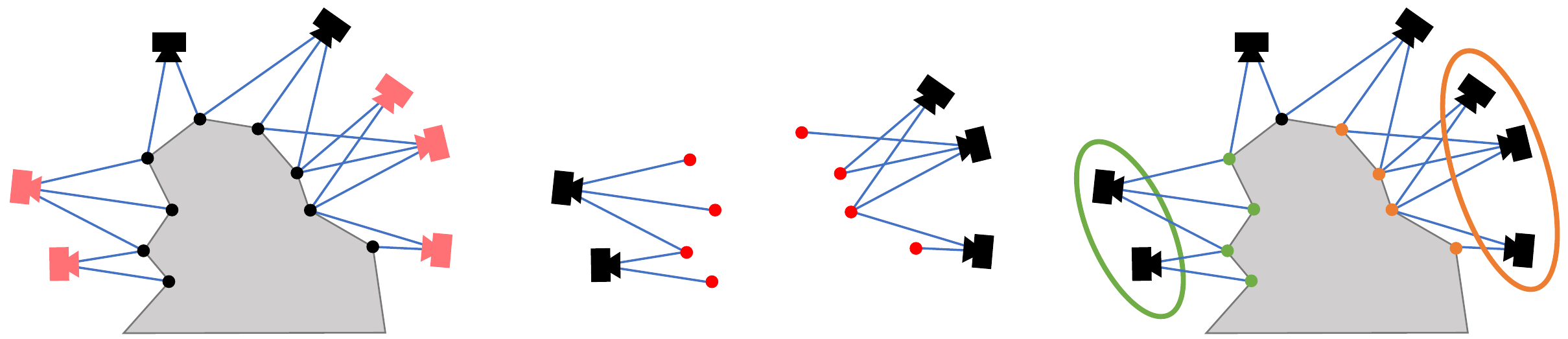}
    \caption{\textbf{Example of covisibility clustering.} 1) Five prior frames (red) are retrieved among all the map keyframes. 2) These, along with the 3D points that they observe (red), correspond to the vertices of a bipartite graph, whose edges (blue) correspond to the 2D-3D observations. 3) Two connected components are found (green and orange). The query image is subsequently matched to each set of corresponding points, starting from the orange component as it contains the most keyframes.\label{fig:covisibility}}%
\end{figure}

The prior frames are clustered based on the 3D structure covisibility: two frames correspond to the same place if they observe some 3D points in common. This amounts to finding the connected components in the bipartite graph composed of frames and observed 3D points, and is fast to compute. An example is show in Figure~\ref{fig:covisibility}.

\subsection{Local matching and Localization}
We now iterate over each place, starting from the one containing the most prior frames, and attempt to estimate a pose using a geometric consistency check. We first compute local descriptor matches between the 3D points that it contains and keypoints in the query image. The number of 3D points considered is now significantly smaller than the total number of points in the map, enabling the use of high-dimensional non-binary descriptors for robust matching with a tractable run-time.

The geometrical consistency of the resulting 2D-3D matches is then assessed by solving the perspective-$n$-point (P$n$P) problem. This amounts to finding the global 6-DoF pose of the camera from $n$ corresponding 3D points in a RANSAC outlier rejection scheme. If a valid pose is found, the process terminates and the query image is successfully localized. Otherwise, it iterates to the next place, until a valid pose is found or until all places have been examined.

%===============================================================================
\section{Experiments and results}
\label{sec:exp}

\subsection{Datasets}
To ensure an optimal generalization of the student retrieval network, it should be trained on a large dataset covering all environments in which it will be deployed. We select the Google Landmarks dataset~\citep{delf}, as it contains a large number of images of landmarks around the world in various scales and illumination conditions, in both urban and non-urban environments. We however take a random subset of 100k images only as the student network is shallow and thus easy to train.

A dataset suitable for the evaluation of the retrieval system should contain a large set of images captured at the same geographical locations in perceptually changing conditions, e.g.\ at multiple times of the day and in diverse weather conditions. An accurate ground truth pose for every image is also necessary. The NCLT dataset~\citep{nclt} is one of the rare datasets to meet these requirements, as it is composed of multiple overlapping outdoor sequences recorded over a timespan of over a year, thus exhibiting large illumination and seasonal changes. We extract images at interval of 5 meters within sequences and define as ground truth global matches pairs of images that have overlapping fields of view, i.e.\ distance and orientation errors smaller than 5 meters and 90 degrees respectively.

As evaluating the full localization pipeline also requires an associated map with 3D points, we record two overlapping 1km-long visual-inertial sequences in the city of Zurich using Google Tango devices. The dataset is recorded at three weeks of interval in different weather conditions and times of the day, making it challenging for current localization methods. The sequences are jointly optimized with the visual-inertial SLAM framework Maplab~\citep{maplab} to produce accurate and aligned maps, each containing approximately 1,400 keyframes and 455k 3D points. Our work does not require expensive Lidar or depth sensor data to build the map.

\subsection{Implementation}
The encoder of MobileNetVLAD is a MobileNet model with depth multiplier of 0.35 and initialized with weights pre-trained on ImageNet~\citep{imagenet}. We predict global descriptors from the second last feature map as it contains fewer channels (320) and thus requires less computation. The VLAD layer has 32 clusters, resulting in a 10,240-dimensional descriptor, which is further downprojected to 4,096. The target descriptors are computed using the NetVLAD implementation of~\citep{netvlad-tf} on images resized to $640\times480$. The images fed to the student are also resized, but additionally converted to grayscale, as most robotics systems do not provide RGB data.

The retrieval system is deployed as a C++ Catkin package~\citep{ros}, making it easy to integrate in existing robotic systems, and performs the nearest neighbors search using the libnabo k-d tree library~\citep{libnabo}. Our localization pipeline is directly integrated into Maplab and can be used in both online and offline modes with existing maps. We employ the P3P-RANSAC implementation~\citep{kneip_cvpr11} available in Maplab with default parameters and use SIFT~\citep{sift} as local descriptor, which is matched using a k-d tree with approximate search precision $\epsilon=3$.

\subsection{Evaluation}

\subsubsection{Retrieval System}
The retrieval system is first evaluated on NCLT with one sequence (2012-01-08) as the index reference and two sequences (2013-02-23 and 2012-08-20) as queries. These cover both cloudy and sunny conditions, with and without snow, and are sufficiently distant in time to exhibit structural changes. Images with no ground truth matches are removed. A query image is considered correctly localized if at least one of the retrieved images contains a correct global match. This is a standard procedure in image retrieval research~\citep{netvlad} and is particularly appropriate to our task as a single correct prior frame can contain sufficient local matches to result in a successful localization. We also evaluate on our Zurich sequences with the same procedure.

\begin{figure}[!htb]
    \centering
    \captionsetup{justification=justified}
    \begin{minipage}{0.48\linewidth}
        \setlength{\tabcolsep}{5pt}
        \begin{tabular}{c|c}
            \toprule
            \makecell{Model} & \makecell{Inference time (ms)}\\ 
            \midrule
            NetVLAD & 2076 \\
            Resnet50+pooling & 141 \\
            MobileNetVLAD & \textbf{55} \\
            \bottomrule
        \end{tabular}  
    \end{minipage}%
    \hspace{1mm}%
    \begin{minipage}{0.4\linewidth}
        \includegraphics[width=\linewidth]{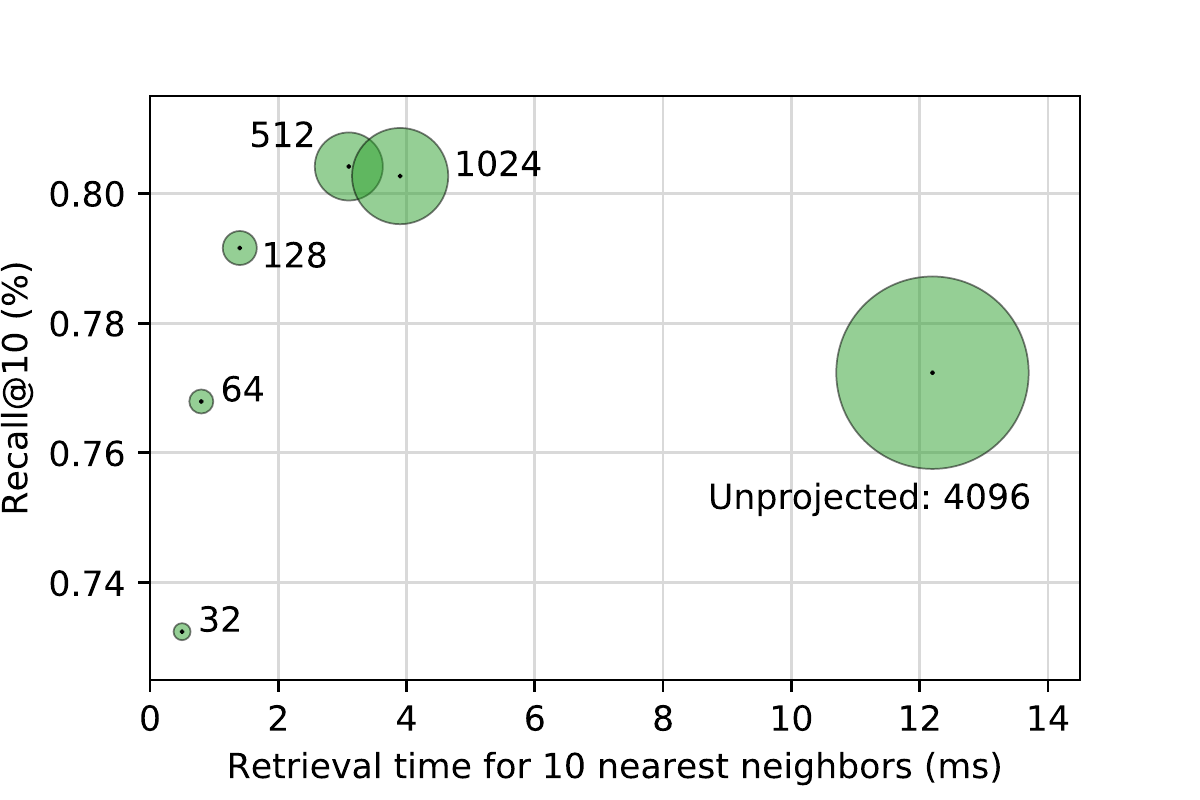}
    \end{minipage}%
  \caption{\textbf{Retrieval timings on a mobile plateform.} We show the inference time on an NVIDIA Jetson TX2 board with embedded GPU (left) and the relationship between the recall of MobileNetVLAD on NCLT and the nearest neighbor search time for several PCA dimensionalities~(right). Our model runs~38~times faster than NetVLAD and significantly benefits from PCA, with a tractable search time.\label{fig:eval:retrieval:timings}}%
\end{figure}

\begin{figure}[!htb]
\centering
\captionsetup{justification=justified}

\begin{subfigure}[b]{0.45\linewidth}
  \centering
  \def\iwidth{0.48}
  \def\cwidth{0.05}

%  \begin{minipage}{\cwidth\textwidth}
%    \rotatebox[origin=c]{90}{\scriptsize{Query 1}}
%  \end{minipage}%
  \begin{minipage}{\iwidth\linewidth}
    \includegraphics[width=\linewidth]{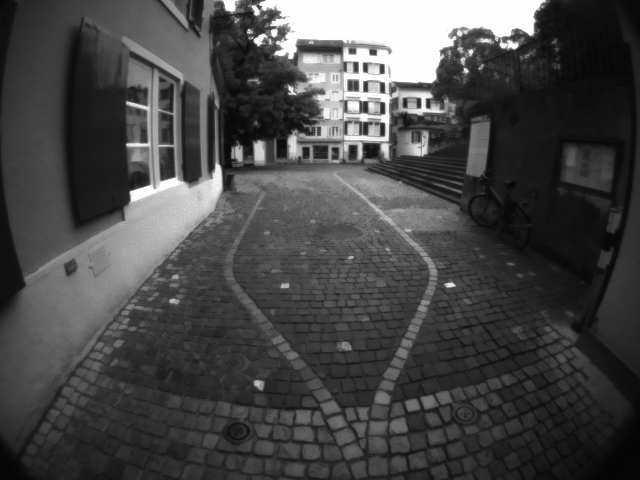}
  \end{minipage}%
  \hspace{1mm}%
  \begin{minipage}{\iwidth\linewidth}
    \includegraphics[width=\linewidth]{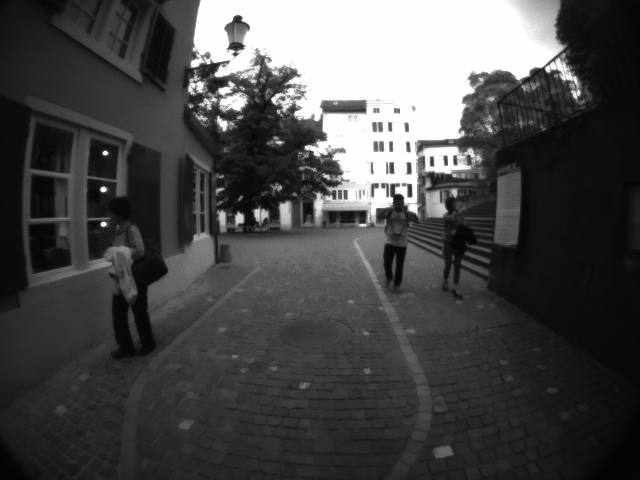}
  \end{minipage}%
%  \begin{minipage}{\cwidth\linewidth}
%    \hfill
%  \end{minipage}%
    \vspace{1mm}
%  \begin{minipage}{\cwidth\textwidth}
%    \rotatebox[origin=c]{90}{\scriptsize{Query 2}}
%  \end{minipage}%
  \begin{minipage}{\iwidth\linewidth}
    \includegraphics[width=\linewidth]{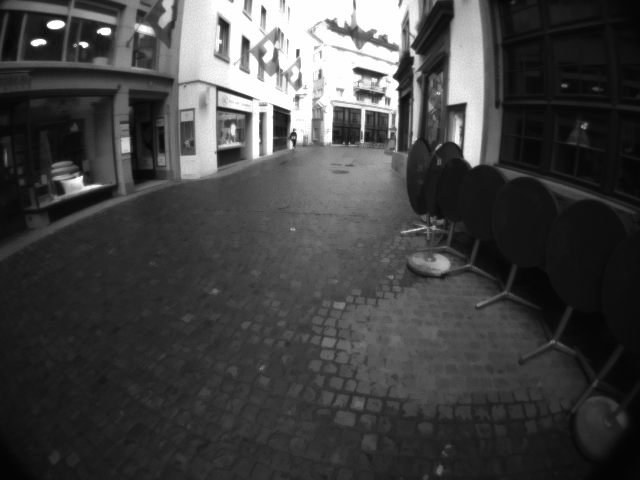}
  \end{minipage}%
  \hspace{1mm}%
  \begin{minipage}{\iwidth\linewidth}
    \includegraphics[width=\linewidth]{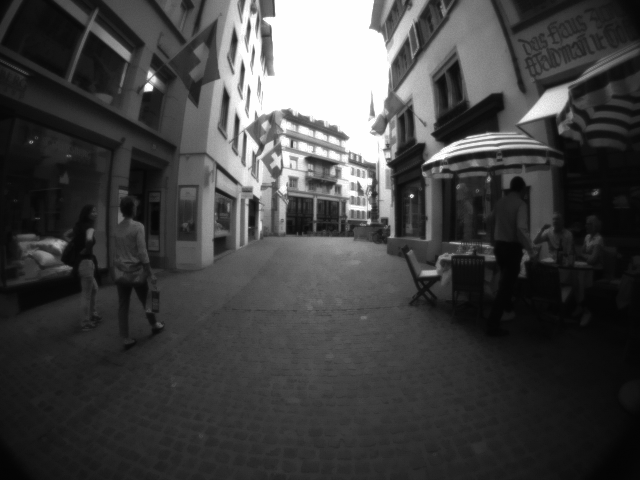}
  \end{minipage}%
%  \begin{minipage}{\cwidth\textwidth}
%    \hfill
% \end{minipage}%
  
  %\vspace{1mm}
  %\begin{minipage}{\cwidth\linewidth}
  %  \hfill
  %\end{minipage}%
  %\begin{minipage}{\iwidth\linewidth}
  %  \centering
  %  \scriptsize{Query image}
  %\end{minipage}%
  %\hspace{1mm}%
  %\begin{minipage}{\iwidth\linewidth}
  %  \centering
  %  \scriptsize{Correctly retrieved image}
  %\end{minipage}%
  %\begin{minipage}{\cwidth\linewidth}
  %  \hfill
  %\end{minipage}%
  
  %\vspace{0.22cm}
  \caption{\textbf{Global retrieval:} queries (left) and correctly retrieved images (right). MobileNetVLAD generalizes well to a new city despite illumination, viewpoint, and structural changes.}%
  \label{fig:eval:samples}%
\end{subfigure}\hspace{4mm}%%
\begin{subfigure}[b]{0.45\linewidth}
  \centering
    \def\iwidth{0.975}
    \includegraphics[width=\iwidth\linewidth]{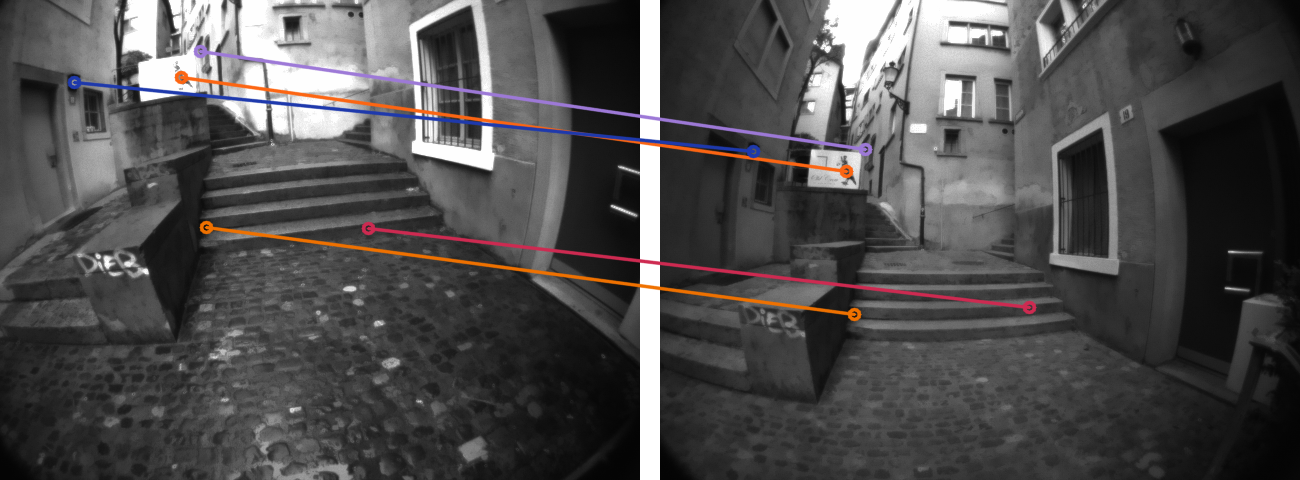}%
    \vspace{1mm}
    \includegraphics[width=\iwidth\linewidth]{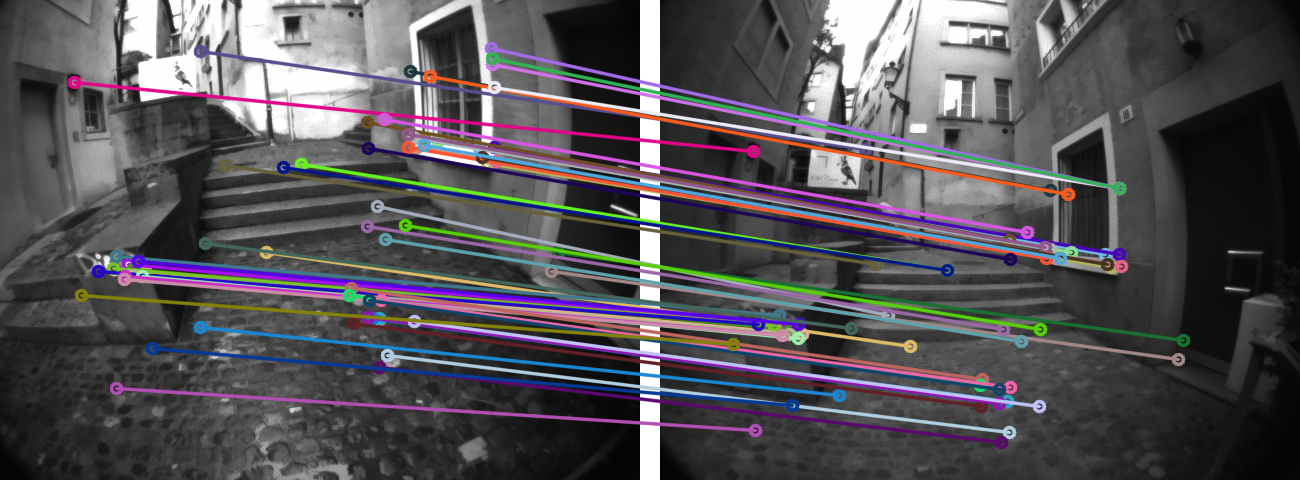}%
%    \vspace{0.21cm}
  \caption{\textbf{Local matching:} query (left) and one map keyframe with matches (right). Direct matching with FREAK (top) fails because of the lack of matches while our method (bottom) succeeds.}
  \label{fig:eval:matches}
\end{subfigure}%%
\vspace{-0.5ex}
\caption{\textbf{Visual samples} of global (\subref{fig:eval:samples}) and local (\subref{fig:eval:matches}) matches on our Zurich dataset.}%
\end{figure}

We compare MobileNetVLAD against the original NetVLAD model trained on the Pitts250k dataset~\citep{pitts250k} as well as a simple baseline consisting of the ResNet50 encoder from DELF~\citep{delf} followed by a global max-pooling head. 
Figures~\ref{fig:eval:recall} and~\ref{fig:eval:retrieval:timings} assess the recall and the timings, respectively. Reducing the descriptor dimensionality with PCA accelerates the global search and can increase the recall by balancing the differences between the descriptor distributions of the training and evaluation environments. MobileNetVLAD performs similarly to NetVLAD but runs much faster, and is thus perfectly suitable for deployment on a mobile platform with real-time constraints. Samples of successful queries are shown in Figure~\ref{fig:eval:samples}.

\definecolor{_red}{RGB}{212,42,47}
\definecolor{_orange}{RGB}{253,127,40}
\definecolor{_green}{RGB}{51,159,52}
\definecolor{_blue}{RGB}{38,120,178}
\definecolor{_purple}{RGB}{147,106,187}

\begin{figure}[!htb]
    \centering
    \captionsetup{justification=justified}
    \begin{subfigure}[c]{0.49\linewidth}
        \centering
        \hspace*{-0.05\textwidth}%
        \includegraphics[width=0.95\textwidth]{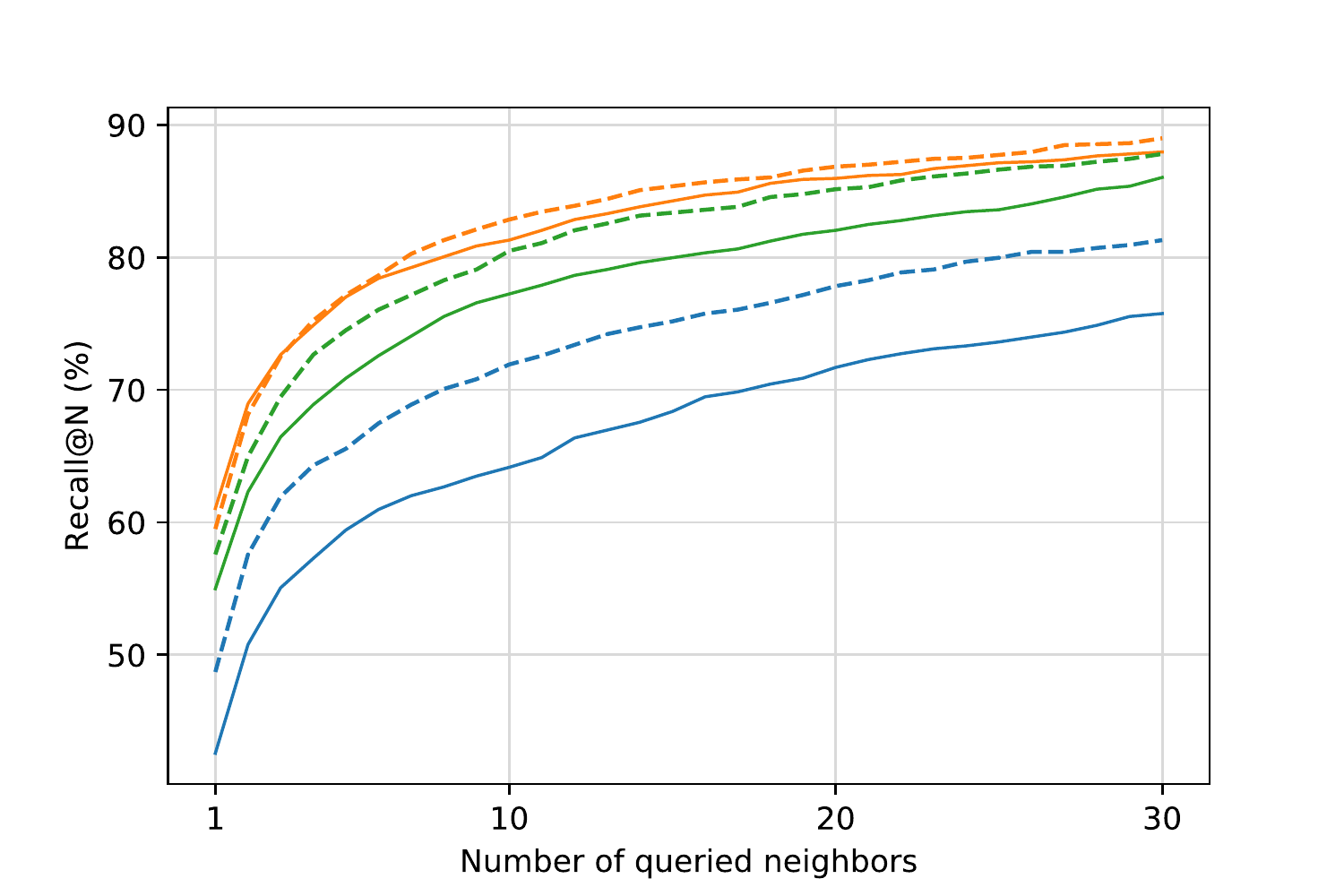}%
        \caption{NCLT dataset}%
    \end{subfigure}%%
    \hspace{2mm}%
    \begin{subfigure}[c]{0.49\linewidth}
        \centering
        \includegraphics[width=0.95\textwidth]{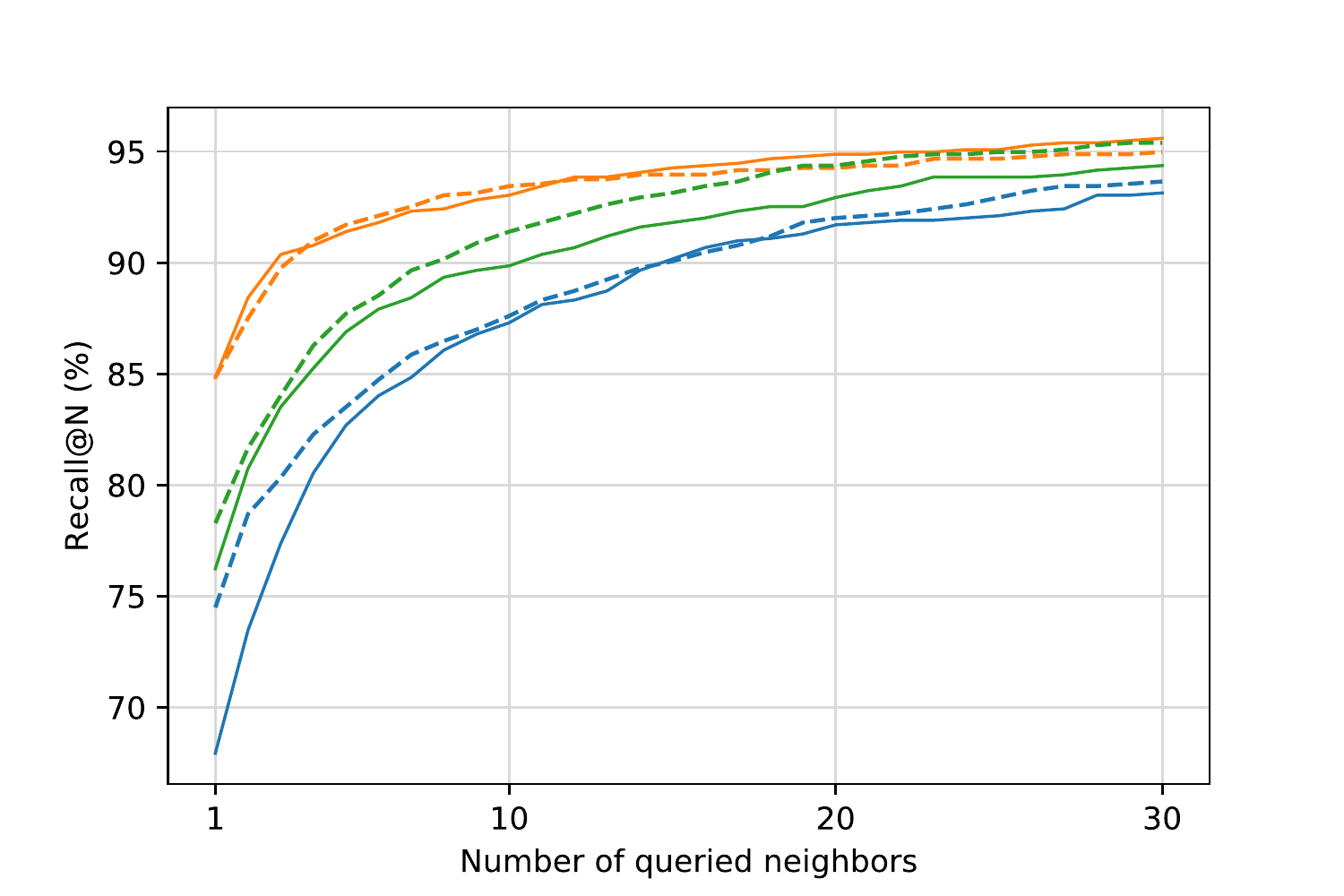}
	\caption{Zurich dataset}%
    \end{subfigure}%%
    \caption{\textbf{Retrieval recall.} Our model {\color{_green}{MobileNetVLAD}} outperforms the {\color{_blue}{Resnet50+pooling}} baseline and approaches the performance of the {\color{_orange}{original NetVLAD}}. Distillation and PCA (dimensionality 512, dashed lines) do not significantly impair the performance.\label{fig:eval:recall}}%.
\end{figure}%
\vspace{-0.3cm}

\vspace{-0.2cm}
\vspace{1ex}

\subsubsection{Pose Estimation}
We proceed to the evaluation of the full localization pipeline on our Zurich dataset. We select one map as the reference localization map and pre-process it offline by computing and indexing the global descriptor of each keyframe and the local descriptors of its 2D keypoints. The query sequence corresponds to the keyframes of the second map, for which we have accurate ground truth poses with respect to the localization map. 

We consider as baseline the direct matching method used in Maplab (Direct+FREAK), which produces state-of-the-art results for resource-constrained systems. It is based on projected~\citep{learned-proj} FREAK binary descriptors~\citep{freak} and performs the search over all the map 3D points using an efficient inverted multi-index~\citep{get-out-of-my-lab}. Covisibility filtering is applied to keep the matches pointing only to the most widely-matched parts of the map. We additionally evaluate the pose estimation sub-system given perfect global prior, computed as the set of frame that have similar orientation and position as the query image. This optimistic baseline (Perfect+SIFT) is equivalent to the LocalSfM of \citet{loc-benchmark}, which was shown to outperform other popular approaches, making further comparisons unnecessary. Under the light of the previous section, and unless otherwise specified, we now use MobileNetVLAD (MNV) with 10 prior frames and reduce the global descriptors dimensionality to 512. We compare both FREAK and SIFT local descriptors (MNV+FREAK and MNV+SIFT respectively).

\begin{figure}[!htb]
    \centering
    \captionsetup{justification=justified}
    
    \begin{minipage}{0.625\linewidth}
        \includegraphics[width=0.98\linewidth]{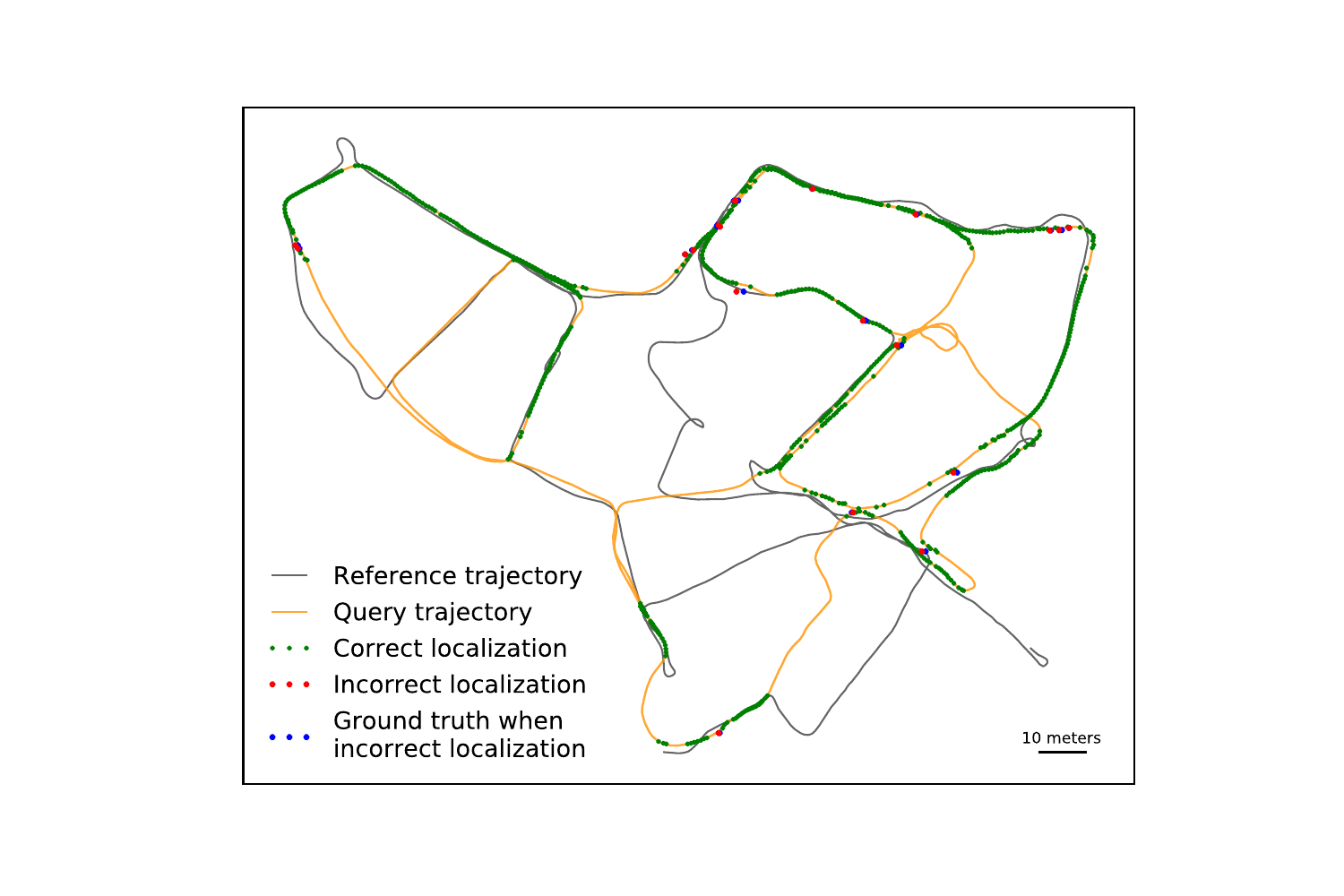}
    \end{minipage}%
    \hspace{1mm}%
    \begin{minipage}{0.365\linewidth}
        \includegraphics[width=\linewidth]{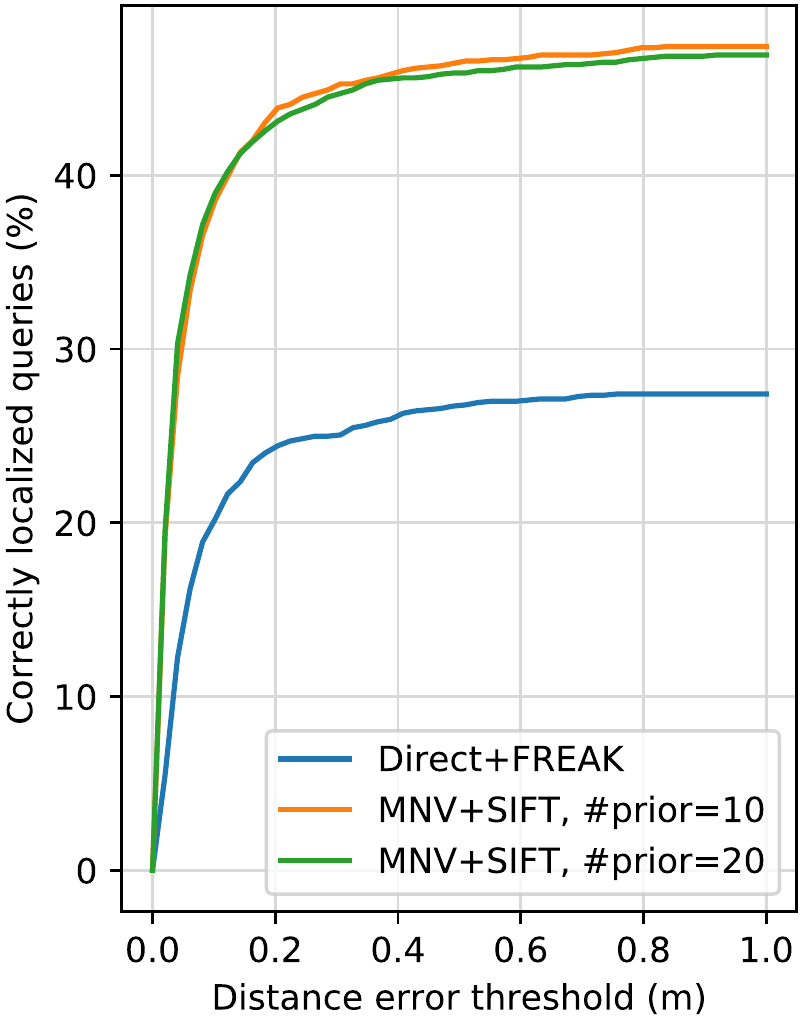}
    \end{minipage}%
    \caption{\textbf{Localization results on our Zurich dataset.} Left: localization map of MobileNetVLAD+SIFT. The reference and query trajectories do not completely overlap, therefore only 69\% of the queries are in fact localizable. Right: cumulative distribution of position errors. MNV+SIFT and Direct+FREAK localize 47.4\% and 27.4\% of the queries, respectively.\label{fig:localization_map}}%
\end{figure}

We first provide some statistics in Table~\ref{tab:eval:loc:stats} and motivate some of our design choices. We then show in Table~\ref{tab:eval:loc} the recall, precision, and median position error of our method and the above-described baselines. Our localization system outperforms the tractable baselines on all three metrics, improving the recall by 18.4\% compared to the state-of-the-art. Surprisingly, FREAK descriptors do not benefit from the global prior. We attribute this to the fact that their correct 2D-3D matches are distributed over more 3D points than the ones observed by the few prior frames. On the contrary, SIFT can easily yield a successful estimate with points observed by a single frame. A typical example of failure of Direct+FREAK is shown in Figure~\ref{fig:eval:matches}. Additional results are shown in Figure~\ref{fig:localization_map}.

Our method performs similarly to the optimistic upper bound, thus confirming that current state-of-the-art deep retrieval systems are accurate enough for environments of such scale. This also reveals that the accuracy of our system is mainly limited by the performance of the local descriptors and their matching, and thus sets the direction for future research.

\begin{table}[!htb]
  \centering
  \captionsetup{justification=justified}
  
  \begin{tabular}{c|*3c}
    \toprule
    Direct+FREAK~\citep{get-out-of-my-lab} & \multicolumn{3}{c}{MNV+SIFT} \\
    \midrule
    \makecell{Number of inlier\\local matches} 
    & \makecell{Number of inlier\\local matches}
    & \makecell{Total number \\of retrieved places} 
    & \makecell{Number of places\\evaluated until success} \\
    \midrule
    25.7 & 45.6 & 4.45 & 1.23 \\
    \bottomrule
  \end{tabular}%
  \vspace{1ex}
  \caption{\textbf{Average statistics over all queries.} Our intermediate retrieval allows to efficiently use SIFT, which produces more correct matches than FREAK, thus increasing the pose accuracy. While 4.45 places are retrieved on average, our algorithm sorts them by decreasing size and thus only evaluates 1.23 of them. This significantly reduces the matching run-time.\label{tab:eval:loc:stats}}%
\end{table}%
\vspace{-0.4cm}
\begin{table}[!htb]
  \centering
  \captionsetup{justification=justified}
  \begin{tabular}{c|*3c|c}
    \toprule
    Method & Direct+FREAK~\citep{get-out-of-my-lab} & MNV+FREAK & MNV+SIFT & Perfect+SIFT\\
    \midrule
    Recall@0.1m (\%)      & 20.1  & 13.9  & \textbf{38.5}  & 39.3 \\
    Precision@0.1m (\%)   & 71.1  & 68.6  & \textbf{80.5}  & 83.3 \\
    Median error (m) & 0.048 & 0.060 & \textbf{0.029} & 0.028 \\
    \bottomrule
  \end{tabular}
  \vspace{1ex}
  \caption{\textbf{Performance of the full localization pipeline.} Recall and Precision are computed for a 0.1m position error. Our system MNV+SIFT significantly outperforms the state-of-the-art FREAK-based direct matching of Maplab (Direct+FREAK) and is very close to the optimistic baseline that assumes perfect retrieval (Perfect+SIFT), thus providing a tractable centimeter-precise pose.\label{tab:eval:loc}}%
\end{table}%
\vspace{-0.5cm}

\subsubsection{Localization Timings}

We measure the run-time of our localization pipeline on an NVIDIA Jetson TX2 board, a mobile platform with embedded GPU widely used for mobile robotics. We report the results in Tables~\ref{tab:eval:timings} and~\ref{tab:eval:timings-params}. Our method runs in real-time and can easily be adjusted to match given run-time constraints.

\vspace*{-0.3cm}
\begin{table}[!htb]
  \centering
  \captionsetup{justification=justified}
  \setlength{\tabcolsep}{5pt}
  \begin{tabular}{*6c}
      \toprule
      \makecell{Network inference} & 
        \makecell{Global search} & 
        \makecell{SIFT Compute} & 
        \makecell{2D-3D matching} & 
        \makecell{PnP-RANSAC} & Total \\
      \midrule
      55 & 2 & 77 & 210 & 32 & \textbf{451} \\
      \bottomrule
  \end{tabular}
  \vspace{1ex}
  \caption{\textbf{Run-time (ms) of the localization pipeline on a mobile platform.} Our system runs online at 2.2 FPS on a Jetson TX2. Only the network inference runs on GPU. The local matching is the bottleneck and can be sped up at the expense of accuracy.\label{tab:eval:timings}}%
\end{table}
\vspace{-0.8cm}
\begin{table}[!htb]
  \centering
  \captionsetup{justification=justified}
  \begin{tabular}{c|ccc|ccc}
    \toprule
    Number of prior frames $N$ & $N=5$ & $N=10$ & $N=20$ & \multicolumn{3}{c}{$N=10$}\\
    Local matching approximation $\epsilon$ & & $\epsilon = 3$ & &  $\epsilon = 9$ & $\epsilon = 3$ & $\epsilon = 0.5$\\
    \midrule
    Recall@0.1m (\%)    & 38.4  & 38.5  & 38.8 & 32.9 & 38.5 & 38.6 \\
    Median error (m)    & 0.031 & 0.029 & 0.027 & 0.035 & 0.029 & 0.029 \\
    Total run-time (ms) & 278 & 451 & 521 & \textbf{206} & 451 & 910 \\
    \bottomrule
  \end{tabular}
  \vspace{1ex}
  \caption{\textbf{Relationship between key parameters and timing.} Two parameters that have a significant impact on the run-time are the number $N$ of prior frames and the approximation factor $\epsilon$ of the local matching k-d tree, whose higher value yields a more approximate search. Decreasing $N$ reduces the run-time and only marginally impairs the accuracy. Increasing $\epsilon$ further accelerates the localization but has a larger negative impact on the performance. There is thus a clear trade-off between accuracy and run-time. Even at 4.9 FPS, we still outperform the current methods by a large margin.\label{tab:eval:timings-params}}%
\end{table}
\vspace{-0.2cm}

%===============================================================================
\section{Conclusion}
\label{sec:conclusion}
This work introduced a new hierarchical localization system that leverages deep learned global image descriptors to first perform a coarse search at the map level, and subsequently estimates a pose using expensive local descriptors. As opposed to existing methods, our system estimates centimeter-precise 6-DoF poses in city-block-scale and GPS-denied environments while running in real-time on a mobile platform, and does not require any training data from the target environment. Our pipeline outperforms current state-of-the-art systems by a large margin and delivers real-time performance on an NVIDIA Jetson TX2 board, making it easy to integrate into mobile robotic projects.

We showed that the accuracy of the proposed method is mainly limited by the performance of the local descriptors and their matching, and that improving current retrieval systems will at this scale only give minor benefits. Recent work on learned local descriptors~\citep{superpoint, sips} has shown promising results, and we believe that future research should head towards combining deep learned global and local descriptors for improved accuracy and efficiency.

%===============================================================================

% The maximum paper length is 8 pages excluding references and acknowledgements, and 10 pages including references and acknowledgements

%\clearpage
% The acknowledgments are automatically included only in the final version of the paper.
%\acknowledgments{If a paper is accepted, the final camera-ready version will (and probably should) include acknowledgments. All acknowledgments go at the end of the paper, including thanks to reviewers who gave useful comments, to colleagues who contributed to the ideas, and to funding agencies and corporate sponsors that provided financial support.}

\clearpage
%\acknowledgments{}

%===============================================================================

\bibliography{refs}  % .bib

\end{document}